\documentclass{article}
\usepackage{arxiv}

\usepackage[utf8]{inputenc} % allow utf-8 input
\usepackage[T1]{fontenc}    % use 8-bit T1 fonts
\usepackage[hidelinks]{hyperref}       % hyperlinks
\usepackage{url}            % simple URL typesetting
\usepackage{booktabs}       % professional-quality tables
\usepackage{amsfonts}       % blackboard math symbols
\usepackage{nicefrac}       % compact symbols for 1/2, etc.
\usepackage{microtype}      % microtypography
\usepackage{mathtools}
\usepackage{amsmath}
\usepackage{amssymb}
\usepackage{amsthm}
\usepackage{multirow}
\usepackage{graphicx}
\usepackage{textcomp}
\usepackage{color}
\usepackage{float}
\usepackage{tabularx}

\newcommand\blfootnote[1]{%
  \begingroup
  \renewcommand\thefootnote{}\footnote{#1}%
  \addtocounter{footnote}{-1}%
  \endgroup
}

\title{Exploring layerwise decision making in DNNs}

\author{
  Coenraad~Mouton and Marelie H.~Davel \\
  Faculty of Engineering, North-West University, South Africa \\
  and CAIR, South Africa \\
  \texttt{ moutoncoenraad@gmail.com} \\
}

\begin{document}
\maketitle
\blfootnote{This is a preprint - the final authenticated publication is available online at \url{https://doi.org/10.1007/978-3-030-95070-5_10}}
\begin{abstract}
While deep neural networks (DNNs) have become a standard architecture for many machine learning tasks,  their internal decision-making process and general interpretability is still poorly understood.   Conversely, common decision trees are easily interpretable and theoretically well understood.   We show that by encoding the discrete sample activation values of nodes as a binary representation, we are able to extract a decision tree explaining the classification procedure of each layer in a ReLU-activated multilayer perceptron (MLP).  We then combine these decision trees with existing feature attribution techniques in order to produce an interpretation of each layer of a model.  Finally, we provide an analysis of the generated interpretations, the behaviour of the binary encodings and how these relate to sample groupings created during the training process of the neural network.

\end{abstract}

% keywords can be removed

\section{Introduction}
\label{sec:introduction}

Deep Neural Networks (DNNs) have become the standard, state-of-the-art solution to many machine learning tasks.  However, the underlying decision-making process is still considered hidden (often referred to as a ``black box''), and does not directly offer any human-interpretable insight. To rectify this lack of transparency, several interpretability methods have been developed.  The most popular of these are feature attribution methods, which provide an indication of the relevance of certain input features to the output of the model, such as Deep Learning Important
FeaTures (DeepLIFT)~\cite{deep_lift}, Layer-wise Relevance Propogation (LRP)~\cite{lrp}, and SHapley Additive exPlanations (SHAP)~\cite{shap}.

While certainly useful, these techniques tend to focus on understanding the way in which a machine learning model links inputs to outputs when considering the model as a whole.  We use a different approach, and consider the internal structures created at each individual layer of a DNN.  More specifically, we make use of the node activation values at each layer and encode this as a binary string representing each training sample.  We then use these discrete binary encodings to derive a decision tree from each layer in a ReLU-activated multilayer perceptron (MLP).  As opposed to neural networks, decision trees provide interpretable structures which inherently explain their internal decision making.  By combining the derived decision trees with a simple mean-sample visualization method, we are able to explore the model's internal classification procedure.

The layout of the paper is as follows: In Section \ref{sec:background} we provide background on (1) binary encodings and how they are obtained from each layer in a neural network and (2) existing decision tree induction methods applied to neural networks.  Following this, in Section \ref{sec:nodes_bin_class}, we study the behaviour of the binary encodings by measuring the number of unique and duplicate encodings at several layers for different architectures and data sets.  In Section \ref{sec:dt_induction} we introduce our algorithm for decision tree induction, before measuring the accuracy and size of the derived decision trees.  Finally, in Section \ref{sec:dt_viz} we visualize these decision trees and shed light on the internal classification mechanisms underlying the neural network.

\section{Background}
\label{sec:background}

In this section we first discuss binary encodings and how they are derived from a neural network, before discussing other existing methods of decision tree induction.

\subsection{Nodes as Binary Classifiers}

Davel et al.~\cite{davel_dnns} have shown that in feedforward ReLU-activated neural networks, each node in any given layer can be viewed as a binary classifier, given that a certain node only results in an activation value greater than 0 for a subset of the training samples.  This is due to the fact that for certain samples the node passes through no information, and for others the pre-activation value is passed through as is.  We use the terms ``ON'' to refer to cases where the node output is greater than zero, and ``OFF'' otherwise.  Through the ON and OFF state of each node, the training samples are partitioned into their respective classes, and thus each hidden layer can be viewed as the combination of several cooperating classifiers.

    % \begin{equation}
    % \label{eq:relu_eq}
    % f(x) = max(0,x) = \left\{
    %         \begin{array}{ll}
    %             0 & \quad x \leq 0 \\
    %             x & \quad x > 0
    %         \end{array}
    %     \right.
    % \end{equation}

%Following this, Davel et al.~\cite{davel_dnns} then further explored 
In the same work, the discrete dynamics of ReLU-activated MLPs was further explored, by encoding the activation values of each layer as a string of binary activation values indicating the ON or OFF state of each node for a given sample.  The resulting string is referred to as a ``binary encoding'', or simply ``encoding'', the  terminology we also use.  They then demonstrated that in the shallower layers of the network, virtually each sample in a given training set has its own unique encoding.  In deeper layers however, the encodings become highly class specific, meaning most samples of the same class share an encoding.  This behaviour was shown to be reproducible over many different architectures and two separate data sets.

In Section \ref{sec:unique_encodings_davel} we verify these findings using our own networks and measurements, and then further explore this phenomenon in Section \ref{sec:duplicate_encodings}.  %Following this, in Section \ref{sec:dt_induction}, we use the concept of binary encodings to derive a decision tree classifier at each layer of a large neural network.    

\subsection{Decision Tree Induction}

Several other methods have been developed for extracting decision trees from neural networks, with varying levels of success.  Sato and Tsukimoto~\cite{cred} introduced the CRED (Continuous/discrete Rule Extractor via Decision tree induction) algorithm, which extracts decision trees using the continuous activation values of nodes in a neural network.  However, this algorithm is only applicable to networks consisting of a single hidden layer~\cite{rule_review}.  Zilke et al.~\cite{deepred} extended the CRED algorithm to networks of multiple hidden layers, but due to its computational complexity it is not suitable for tasks with a large number of input features (such as image classification).  

These methods extend the decision tree extraction to the input features, and furthermore consider the neural network in its entirety.  Conversely, we consider each hidden layer as a separate classifier, and opt to relate each layer's decision tree to the input features through a heatmap based method.  This allows each decision tree to remain small in size, and thus easily interpretable, and furthermore allows us to explain each layer individually.

%%%%%%%%%%%%%%%%%%%%%%%%%%%%%%%%%%%%%%
%%%%%%%%%%%%%%%%%%%%%%%%%%%%%%%%%%%%%%
\section{The Behaviour of Binary Encodings}
\label{sec:nodes_bin_class}

In this section we explore the phenomenon of binary encodings, and how they relate to sample-groupings in a neural network.

\subsection{Unique Encodings}
\label{sec:unique_encodings_davel}

\begin{figure}[h]
        \centering
        \includegraphics[width=0.49\linewidth]{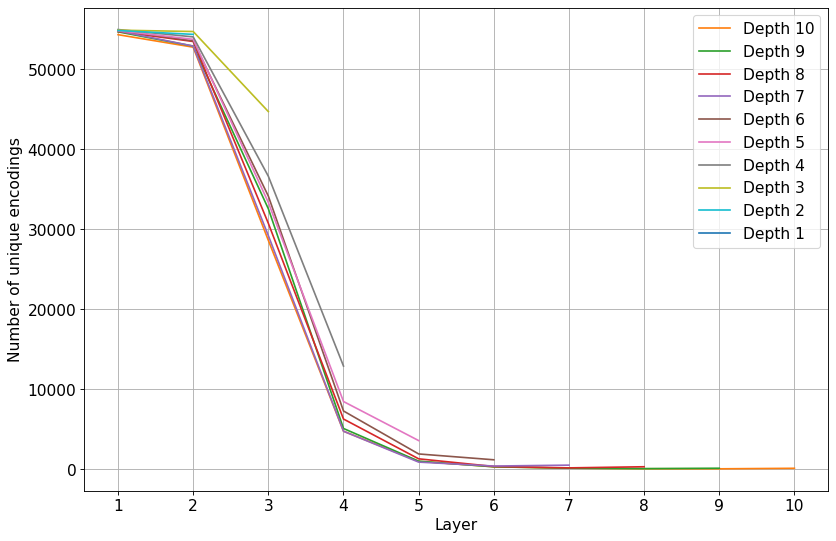}
        \includegraphics[width=0.49\linewidth]{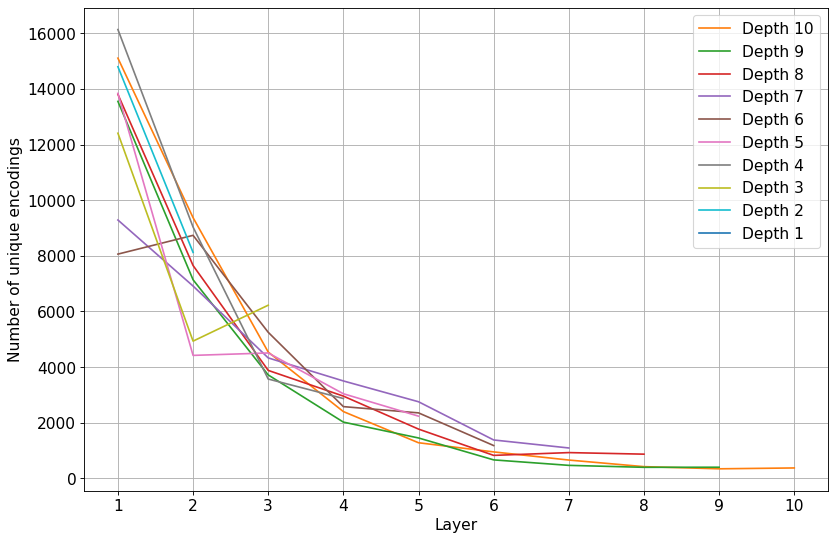} \\
        \includegraphics[width=0.49\linewidth]{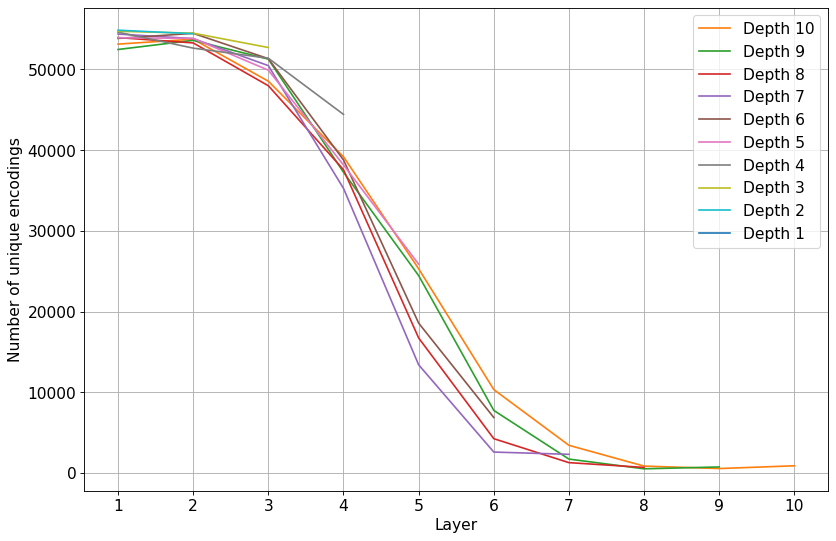}
        \includegraphics[width=0.49\linewidth]{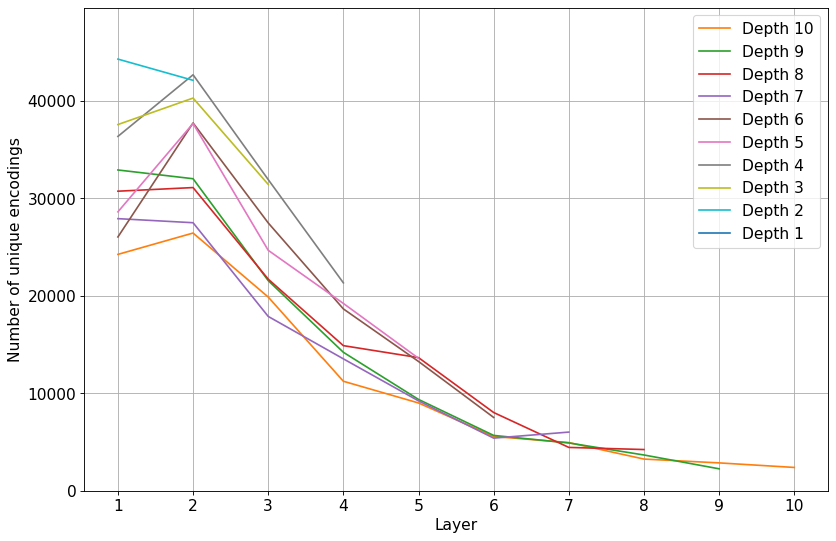}        
        \caption{Number of unique encodings per layer for MNIST and FMNIST networks with 1 to 10 hidden layers. Top left: MNIST-100. Top right: MNIST-20. Bottom left: FMNIST-100. Bottom right: FMNIST-40}
        \label{fig:unique_activation_sequences}
\end{figure}

To confirm the findings of Davel et al.~\cite{davel_dnns}, and provide an understanding of the behaviour of binary encodings at different layers of a network, we devise the following experiment:
  We train 10 ReLU-activated MLPs on the MNIST data set~\cite{MNIST}, with a depth of 1 to 10 hidden layers, where each layer has a width of 100. A bias node is added to the first layer only\footnote{This simplifies theoretical analysis without compromising performance of a ReLU-activated MLP~\cite{davel_dnns}}. We then measure the number of unique binary encodings for all 55,000 training samples at each individual layer for all 10 networks.  We repeat this experiment using a layer width of 20;  and do the same for the Fashion-MNIST data set (FMNIST)~\cite{fashion_mnist} with layer widths of 100 and 40.  The results for all four sets of networks are shown in Figure \ref{fig:unique_activation_sequences}.
  We refer to these networks by their dataset and width, namely: MNIST-100, FMNIST-100, MNIST-20, and FMNIST-40.
 
We use a training set size of 55,000 samples (where a `sample' is a single image), while 5,000 separate samples are used for validation.  Each data set has a separate test set of 10,000 samples.
All networks are trained to 100\% train accuracy. (We do not employ early stopping.)  We use the Adam optimizer~\cite{adam}, with cross-entropy as loss function and perform learning rate decay: every $step\_size$ epochs the learning rate is multiplied by a decay constant $gamma$.
Standard hyperparameter settings are used (batch size of 128, no regularisation), except for the initial learning rate, decay step size, and decay gamma which are optimized on the validation set to ensure interpolation, as shown in Table \ref{tab:hyperparams}, along with the range of test set accuracy for each set of models.

\begin{table}
\centering
\caption{Training hyperparameters and accuracy for MNIST and FMNIST models with 1 to 10 hidden layers and varying widths}
\label{tab:hyperparams}
\resizebox{\columnwidth}{!}{%
\begin{tabular}{c|c|c|c|c|c|c}
\multicolumn{1}{l|}{\textbf{Data set}} & \multicolumn{1}{l|}{\textbf{Width}} & \multicolumn{1}{l|}{\textbf{Learning rate}} & \multicolumn{1}{l|}{\textbf{Decay gamma}} & \multicolumn{1}{l|}{\textbf{Decay step size}} & \multicolumn{1}{l|}{\textbf{Train accuracy (\%)}} & \multicolumn{1}{l}{\textbf{Test accuracy (\%)}} \\ 
\hline
MNIST & 100 & 0.001 & 0.990 & 5 & 100.00 & \textcolor[rgb]{0.129,0.129,0.129}{97.82 to~}\textcolor[rgb]{0.129,0.129,0.129}{98.19} \\
FMNIST & 100 & 0.001 & 0.990 & 5 & 100.00 & \textcolor[rgb]{0.129,0.129,0.129}{87.96 to~}\textcolor[rgb]{0.129,0.129,0.129}{89.10} \\
MNIST & 20 & 0.002 & 0.500 & 100 & 100.00 & \textcolor[rgb]{0.129,0.129,0.129}{94.62 to~}\textcolor[rgb]{0.129,0.129,0.129}{96.12} \\
FMNIST & 40 & 0.002 & 0.850 & 50 & 100.00 & 85.23 to~\textcolor[rgb]{0.129,0.129,0.129}{86.70}
\end{tabular}
}
\end{table}

Observing the MNIST and FMNIST networks with a layer width of 100 in Figure \ref{fig:unique_activation_sequences}, a clear pattern emerges. Initially, in the first layer, nearly every sample has its own unique encoding, as we find the number of unique encodings to be close to the total number of samples in the training set (55,000 samples) for all the networks.  As the depth increases, we observe a sharp decline, indicating that encodings are shared by many samples.  Finally, for the deeper networks, we observe that very few encodings remain in the last few hidden layers, implying that there are only one or two encodings shared by all the samples of a class.  This is concurrent with the observations initially made by Davel et al.~\cite{davel_dnns}.  

For the narrower networks (MNIST-20 and FMNIST-40) the general pattern appears to be similar, except that the initial number of unique encodings in the first layers is much lower (note the difference in scale of the y-axis).  This seems to indicate that by decreasing the width, the network is forced to share encodings among samples, even in the first layer.

In the following section we study the behaviour of these encodings more closely. 

\subsection{Duplicate Encodings}
\label{sec:duplicate_encodings}

Given that the number of encodings decreases as the network depth increases, the implication is that samples share encodings.  We expect that samples of the same class would share an encoding, but samples of different classes less so.

\begin{figure}[h]
        \centering
        \includegraphics[width=0.49\linewidth]{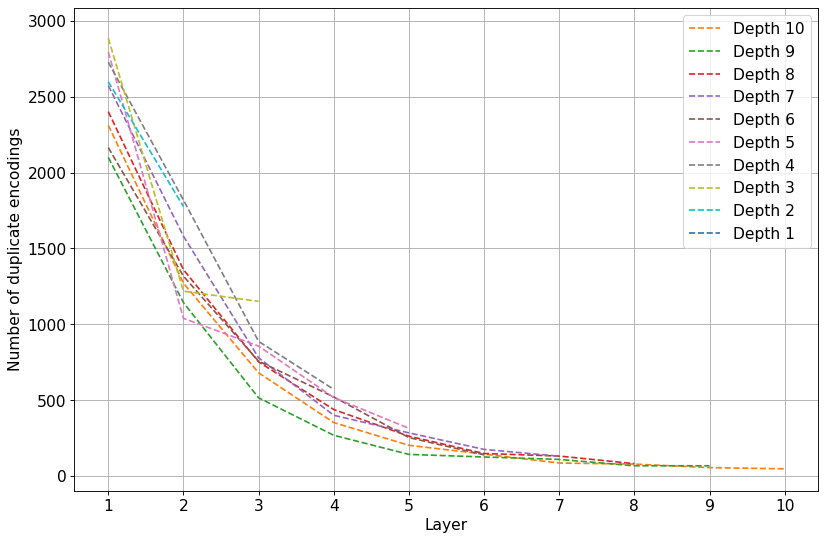} \\
        \includegraphics[width=0.49\linewidth]{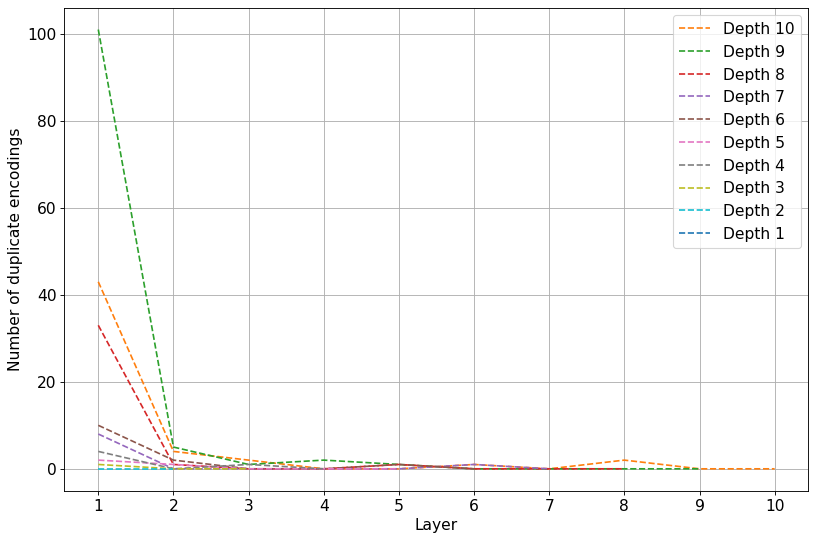} 
        \includegraphics[width=0.49\linewidth]{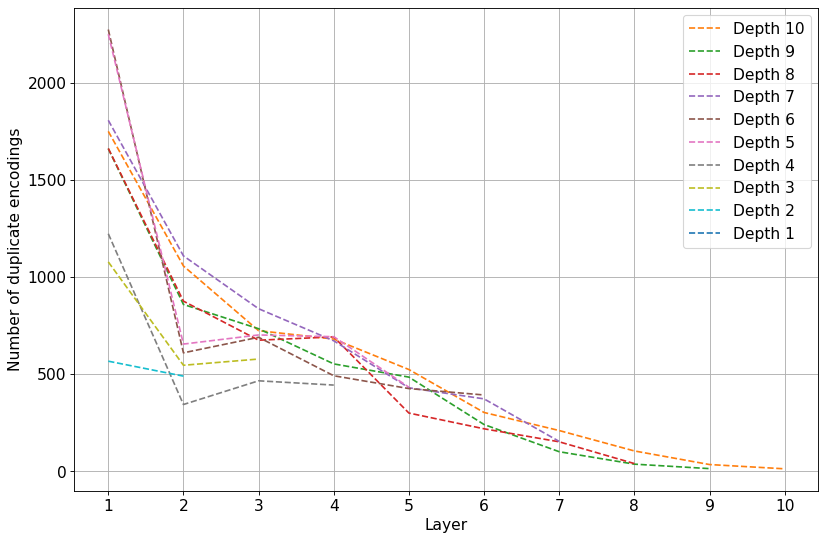}
        \caption{Number of cross-class duplicate encodings per layer for MNIST and FMNIST networks with 1 to 10 hidden layers. Top: MNIST-20. Bottom left: FMNIST-100. Bottom right: FMNIST-40}
        \label{fig:duplicate_activation_sequences}
\end{figure}

We use the term ``duplicate encodings'' to refer to encodings that are shared by samples of {\em more than a single class}.  Using the same networks from the previous section, we measure the number of duplicate encodings at each layer of the network.  For the MNIST networks with a layer width of 100, we find that there are no duplicate encodings at any layer, for any of the 10 networks.  We show the number of duplicate encodings at each layer for the other three sets of networks in Figure \ref{fig:duplicate_activation_sequences}.

In the case of the MNIST-20 networks, we observe that initially there are many duplicate encodings at the first layer, whereafter the number of duplicates steadily decreases.  At the final layers very few remain, indicating that most of the remaining encodings are highly class specific (they only correspond to samples from a single class).  The pattern for the FMNIST-40 networks is very similar, although the decline is slightly less gradual.  Observing the results of Figure \ref{fig:duplicate_activation_sequences} for the wide, FMNIST-100 networks, we see a small number of duplicate encodings at the first layer (once again, note the difference in the scale of the y-axis), and then a rapid decline to almost zero duplicates for hidden layers 3 through 10.  This result, in conjunction with the zero duplicate encodings for the MNIST-100 networks, suggest that networks that are sufficiently wide do not share many encodings between different samples of different classes.

While we have identified that duplicate encodings do occur, we now consider how and why they occur.  Through visual inspection of samples of different classes that share an encoding, we identify 3 distinct scenarios that cause duplicate encodings to occur in the narrow MNIST and FMNIST networks.  We use the term ``idiosyncratic'' to refer to samples that are visually very dissimilar to their class-majority, and ``typical'' for samples that are visually similar to the majority of samples from that class.  The different cases are listed below, and an example of each case for both data sets is shown in Table \ref{tab:duplicate_examples}.  All examples are found at the fifth hidden layer of a MNIST-20 and FMNIST-40 network with 10 hidden layers.  

\begin{enumerate}
    \item Idiosyncratic-to-typical: This occurs when a single, or very few (generally less than 20), idiosyncratic sample(s) share an encoding with many other, more typical samples from a different class.  This can further be split into two sub-cases:
    \begin{enumerate}
        \item Idiosyncratic-to-many: In this case, the few idiosyncratic samples from one class shares an encoding with a very large number of samples from another class.  We show two examples of this in the first row of Table \ref{tab:duplicate_examples}.  For MNIST, we find that 4 idiosyncratic samples from class 5 share an encoding with 3,221 samples from class 3.  In the case of FMNIST, we show how a single idiosyncratic sample from the ``shirt'' class shares an encoding with 2,189 other samples from the ``trouser'' class.
        \item Idiosyncratic-to-few: A duplicate encoding is shared between a few idiosyncratic samples of one class and a small subgroup of samples of another class.  In the second row of Table \ref{tab:duplicate_examples} we show how a strange sample from class 7 shares an encoding with 45 samples from class 1 for MNIST.  Similarly, for FMNIST we show an encoding that is shared by one sample from the ``coat'' class and 53 others of the ``shirt'' class.    
    \end{enumerate}
    \item Idiosyncratic-to-idiosyncratic: This occurs when two small groups of idiosyncratic samples of different classes share an encoding.  The third row of Table \ref{tab:duplicate_examples} shows four idiosyncratic MNIST samples, two from class 2 and two other from class 3 that share an encoding.  For FMNIST we show an encoding that is shared by three different classes, namely two of ``t-shirt'' and two others of ``shirt'' and ``dress'' respectively.  We do note, however, that this also often occurs for slightly larger groups (once again, generally less than 20 samples in size each) of idiosyncratic samples.
\end{enumerate}

While we have listed the most significant cases that we have observed, many cases of shared encodings are not always as easily distinguishable as the examples shown in Table \ref{tab:duplicate_examples}.  While some samples are certainly more typical than others, we believe that there is a spectrum between the two extremes of typical and idiosyncratic samples.  Furthermore, we believe that the visual overlap between samples of different classes plays a large role in the formation of duplicate encodings. 

In the following section, we introduce our method for extracting layerwise decision trees from neural networks, and the reason for our investigation into duplicate encodings also becomes clear.

\begin{table}[h]
\centering
\caption{Examples of different cases of duplicate encodings, where a single encoding is shared by samples of more than one class. All examples are found at the fifth hidden layer of a depth-10 MNIST-20 and FMNIST-40 network.}
\label{tab:duplicate_examples}
\centering
\resizebox{\columnwidth}{!}{%
\begin{tabular}{|c|l|l|p{0.4\linewidth}|} 
\hline
\textbf{Case} & \multicolumn{1}{c|}{\textbf{MNIST}} & \multicolumn{1}{c|}{\textbf{FMNIST}} & \multicolumn{1}{c|}{\textbf{Description}} \\ 
\hline
\begin{tabular}[c]{@{}c@{}}Idiosyncratic\\to\\many\end{tabular} & \begin{tabular}[c]{@{}l@{}}\includegraphics[scale=0.35]{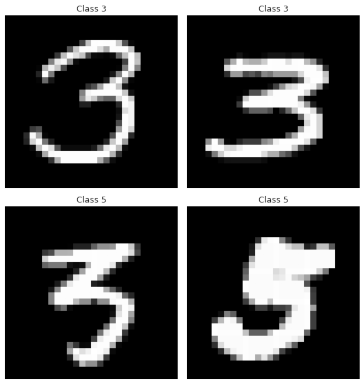}\end{tabular} & \begin{tabular}[c]{@{}l@{}}\includegraphics[scale=0.35]{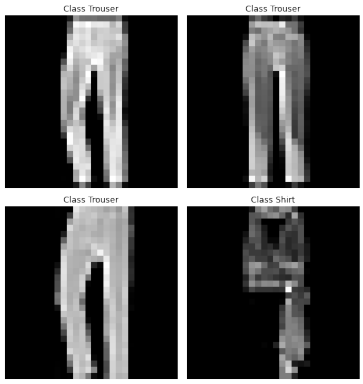}\end{tabular} & 
\begin{tabular}[c]{p{0.95\linewidth}}\textbf{MNIST}: Idiosyncratic samples (4) from class 5 share an encoding with many (3,221) samples of class 3. \\\textbf{FMNIST}: A single idiosyncratic sample of class ``shirt'' shares an encoding with 2,189 samples of class ``trouser''.\end{tabular}\\ 
\hline
\begin{tabular}[c]{@{}c@{}}Idiosyncratic\\to\\few\end{tabular} & \begin{tabular}[c]{@{}l@{}}\includegraphics[scale=0.35]{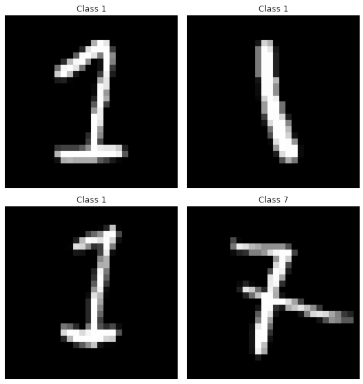}\end{tabular} & \begin{tabular}[c]{@{}l@{}}\includegraphics[scale=0.35]{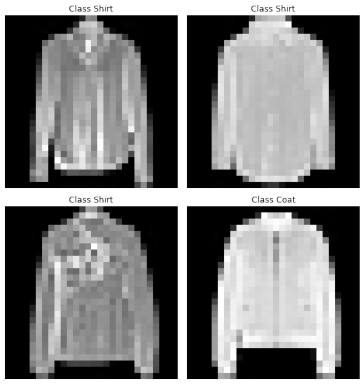}\end{tabular} & 
\begin{tabular}[c]{p{0.95\linewidth}}\textbf{MNIST}: A single idiosyncratic sample of class 7 shares an encoding with 45 samples of class 1.\\\textbf{FMNIST}: A single idiosyncratic sample of class ``coat'' shares an encoding with 53 samples of class ``shirt''.\end{tabular}\\ 
\hline
\begin{tabular}[c]{@{}c@{}}Idiosyncratic\\to\\idiosyncratic\end{tabular} & \begin{tabular}[c]{@{}l@{}}\includegraphics[scale=0.35]{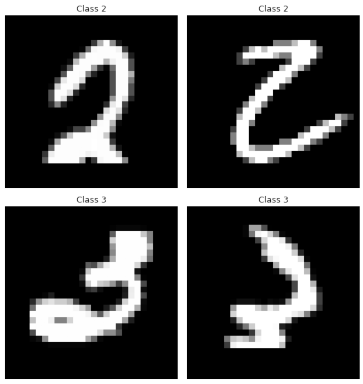}\end{tabular} & \begin{tabular}[c]{@{}l@{}}\includegraphics[scale=0.35]{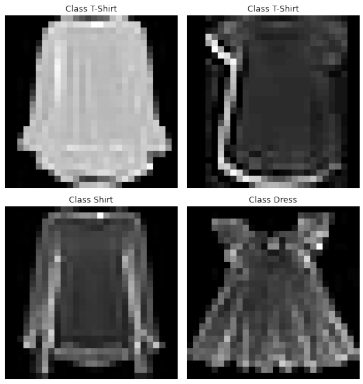}\end{tabular} & 
\begin{tabular}[c]{p{0.95\linewidth}}\textbf{MNIST}: Four idiosyncratic samples from classes 2 and 3 share an encoding. \\\textbf{FMNIST}: Four idiosyncratic samples from the classes ``t-shirt'' (2), ``shirt'' (1), and ``dress'' (1) share an encoding.\end{tabular} \\
\hline
\end{tabular}
}
\end{table}

\section{Layerwise Decision Trees}
\label{sec:dt_induction}

Given the knowledge of binary encodings and their behaviour expressed in the previous section, we now use these encodings to extract a decision tree at each layer of a ReLU-activated MLP, in order to better understand the internal classification procedure of the network. First, in Section \ref{sec:dt_algorithm}, we explain our decision tree extraction algorithm.  In the following section (Section \ref{sec:dt_results}) we show the accuracy of these decision trees at each layer and compare them to that of the neural network from which they are derived.  

\subsection{Algorithm}
\label{sec:dt_algorithm}

We follow a straightforward approach to convert a given layer into a decision tree.  First we find the activation values of each node for each sample at a specific layer.  Following this, we then convert these activation values to binary encodings: a $w$-dimensional vector of 0's and 1's for each sample, with $w$ the width of the layer being probed. Hereafter we identify the unique and duplicate encodings.  We then label each encoding according to the class it corresponds to.  If there is more than one class for the encoding (a duplicate), we label it according to the class of the majority of the samples.  We can then further prune this list of encodings and their corresponding labels by disregarding any that occur for less samples than a given threshold, with the threshold becoming an important hyperparameter.  The remaining list of encodings are then used as the training dataset with which to construct a decision tree. The training set therefore consists of a set of unique binary encodings, each associated with a single class.

This implies that the decision tree algorithm only has access to the ON and OFF states of the nodes as features when building a tree.  Nodes that are ON then provides a 'path' of node states that are required for each class, e.g. if nodes a and b are ON, the sample belongs to class 1, etc.
For decision tree induction, we use the CART (Classification and Regression Tree) algorithm~\cite{cart} and the Gini index~\cite{cart} as splitting criterion.  Due to the fact that some classes might have more unique encodings than others, we treat the data set as balanced, meaning the respective classes are equally weighted.  We do not limit the depth of the decision trees or apply any pruning. Experiments are conducted using the CART implementation from \textit{scikit-learn}\footnote{https://scikit-learn.org/stable/modules/tree.html}. 

To evaluate the accuracy of a decision tree, we pass each sample of the train or test set through the neural network, find its binary encoding at the specific layer that the decision tree represents, and then pass this encoding as input to the decision tree for classification.

\subsection{Results}
\label{sec:dt_results}

\begin{figure}[h]
        \centering
        \includegraphics[width=0.49\linewidth]{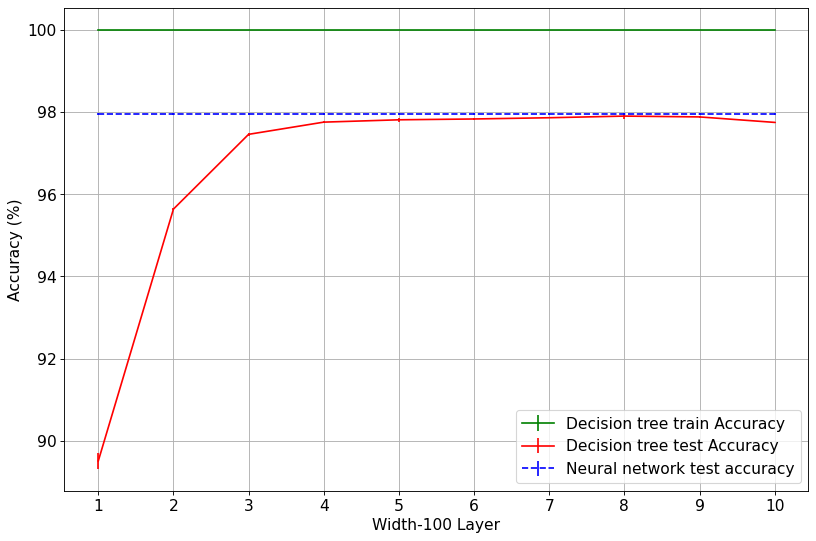}
        \includegraphics[width=0.49\linewidth]{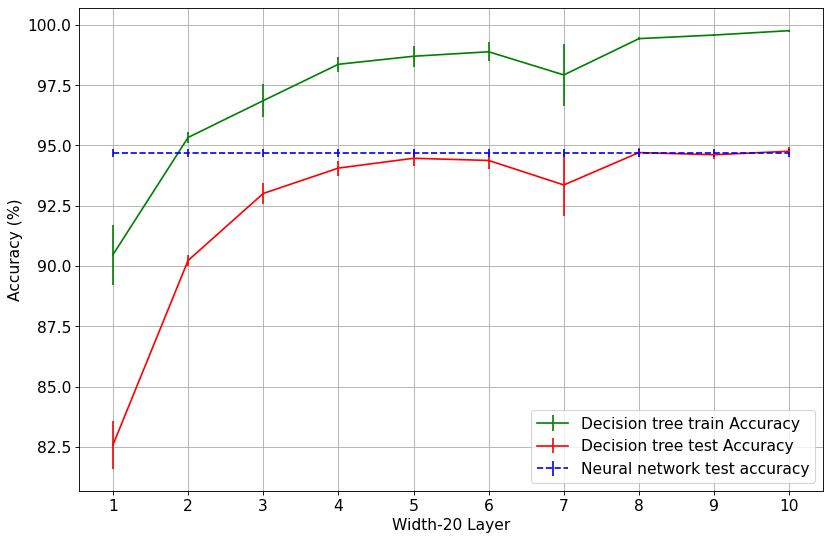} \\
        \includegraphics[width=0.49\linewidth]{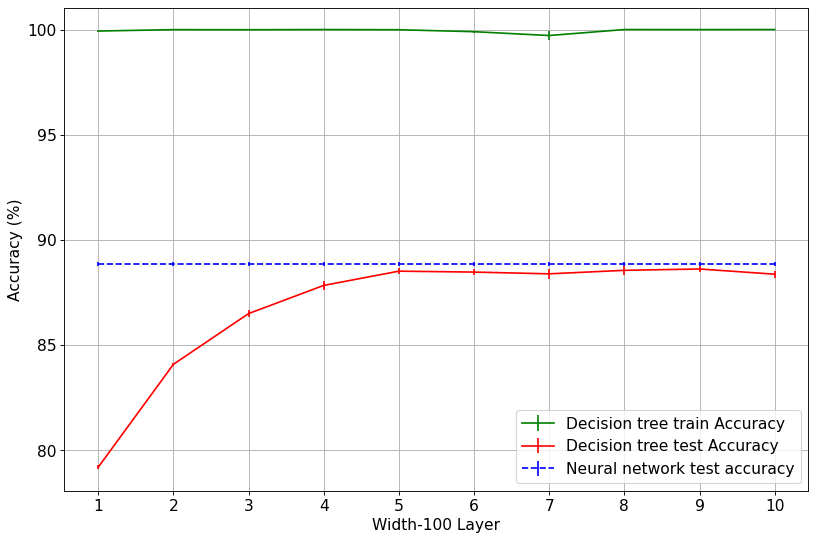}
        \includegraphics[width=0.49\linewidth]{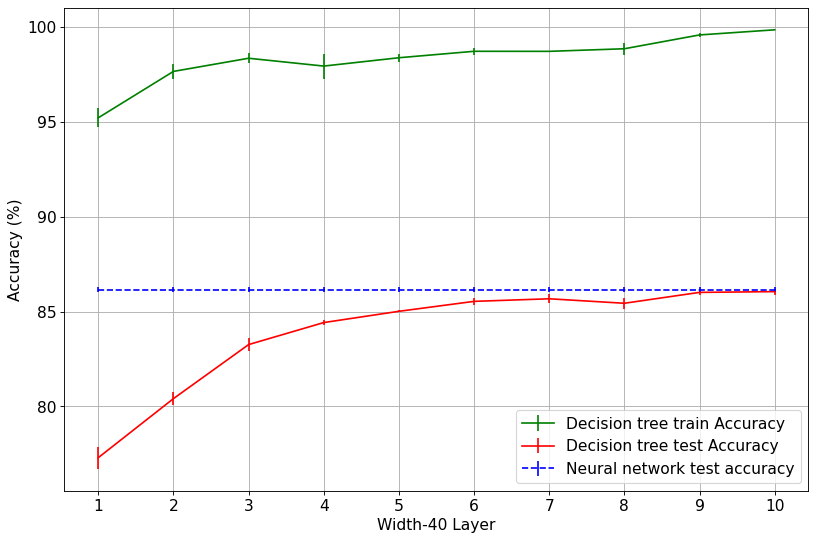}
        \caption{Decision tree train (green) and test (red) accuracy per layer for MNIST and FMNIST networks with 10 hidden layers averaged over three initialization seeds, with a threshold of 0.  The mean test accuracy of the neural network is shown by the horizontal blue line.  Error bars indicate the standard error. Top left: MNIST-100. Top right: MNIST-20. Bottom left: FMNIST-100. Bottom right: FMNIST-40}
        \label{fig:decision_tree_accuracy_per_layer}
\end{figure}

To determine the efficacy of the proposed decision tree extraction algorithm, we use the same setup as in Section \ref{sec:nodes_bin_class}, but generate 3 versions of each network, using 3 different random initialization seeds.  
For each network and seed, we extract a decision tree from each layer, and measure its accuracy on the train and test data sets.  We then take the mean accuracy of the different seeds, and also include error bars indicating the standard error.  We show the average train accuracy (green) and test (red) accuracy for each decision tree at each layer in Figure \ref{fig:decision_tree_accuracy_per_layer}.  We also indicate the average test accuracy of the neural network with a horizontal blue line.  For these results, we use a threshold of 0 when creating the decision trees, meaning no encodings are disregarded from the training data set.

Observing the results in Figure \ref{fig:decision_tree_accuracy_per_layer} for the wide MNIST-100 and FMNIST-100 networks, we see that the decision trees achieve 100\% train accuracy at virtually every layer.  This is due to the small number of duplicate encodings (or 0 duplicates in the case of MNIST), meaning very few samples are unaccounted for by the binary encodings.  Furthermore, the test accuracy of the decision trees in the deeper layers also match the test accuracy of the model.  This implies that the test set samples produce encodings which are very similar to those produced by the train set.  For the narrower MNIST-20 and FMNIST-40 networks, we observe a steady increase in train accuracy as the network depth increases.  Once again, this is due to the number of duplicate encodings decreasing along with the depth of the network.  We also observe that the decision trees in the deeper layers very closely approximate the test accuracy of the network.

\begin{figure}[h]
        \centering
        \includegraphics[scale=0.4]{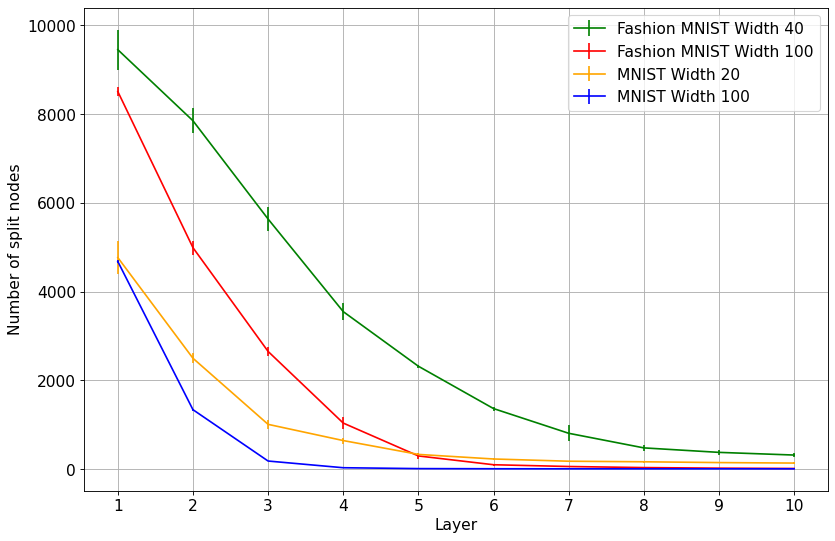}
        \caption{The number of split nodes for each decision tree per layer for MNIST and FMNIST networks with 10 hidden layers averaged over three initialization seeds. Error bars indicate the standard error.} %Green: FMNIST-40. Red: FMNIST-100. Orange: MNIST-20.  Blue: MNIST-100%
        \label{fig:decision_tree_features_per_layer}
\end{figure}

While we find that our decision tree algorithm works well and provides good accuracy, it can only provide a clear interpretation of the layerwise decision making of the neural network if it is sufficiently concise.  We measure the size of each decision tree by finding the number of ``split nodes'', where a split node is a feature used as a splitting criterion in the tree, in other words, each node in the tree that is not a leaf node.  In Figure \ref{fig:decision_tree_features_per_layer} we show the average number of split nodes for each decision tree at each hidden layer.  We once again include error bars indicating the standard error for the three initialization seeds used.

Figure \ref{fig:decision_tree_features_per_layer} shows that the decision trees generated in the shallower layers are exceptionally large, with over 4,000 features for MNIST and over 8,000 for FMNIST in the first layer.  This implies that each node in the layer is re-used many times as a splitting criterion when constructing the decision tree.  However, in the deeper layers we observe that the number of split nodes used by the decision tree is less than the total number of nodes in that specific layer of the neural network.  

Naturally, decision trees that are very large are not easily interpretable, however we can rectify this by applying a threshold when generating the encodings for the decision tree data set (as discussed earlier in Section \ref{sec:dt_algorithm}).    
Making use of the MNIST-100 network, we measure the number of unique encodings, and number of split nodes, after applying a threshold of 0, 100 and 1,000.  This implies that any encoding that does not occur for at least 0, 100 or 1,000 samples respectively are discarded.  Results are shown in Table \ref{tab:encodings_and_features_over_thresholds}, again averaged over three seeds, and rounded to the nearest integer.  We also show the mean train and test accuracy of the decision tree at each layer after applying the specified threshold in Table \ref{tab:accuracy_over_thresholds}.

It is clear that applying a threshold of 100 or 1,000 for the first 2 or 3 layers respectively results in zero encodings remaining in the decision tree's data set.  This is to be expected, as originally observed in Figure \ref{fig:unique_activation_sequences}: very few samples share encodings in the shallow layers of the network.  However, after applying a threshold at the fourth hidden layer the number of encodings is drastically decreased, along with only a slight decrease in accuracy.  This is due to the fact that the majority of samples for each class are contained in very few encodings.  In conjunction with this, we observe a similar decrease in the number of split nodes of each decision tree, implying that each decision tree (once visualized) would provide a more interpretable explanation of the classification procedure for each layer.  Finally, it is quite remarkable that for the final hidden layer (the tenth layer) the number of encodings can be reduced to only 15, and the decision tree still achieves an excellent train and test accuracy (97.97\% and 95.75\% respectively).

In conclusion, the results of Tables \ref{tab:encodings_and_features_over_thresholds} and \ref{tab:accuracy_over_thresholds} show that one can disregard encodings that do not correspond to many samples and significantly decrease the size of each decision tree.  Furthermore, by selecting the correct threshold, the decision tree can be kept concise and still achieve relatively high accuracy on both the train and test data sets.
In the following section we demonstrate a visualization of these decision trees and how they explain the layerwise decision making in a neural network.

\begin{table}[H]
\caption{Average number of encodings and number of split nodes of decision trees extracted at each layer of a MNIST-100 network with different thresholds applied.  Average is obtained over three different initialization seeds.}
\label{tab:encodings_and_features_over_thresholds}
\centering
\begin{tabular}{lcccc}
\multicolumn{1}{c}{\textbf{\begin{tabular}[c]{@{}c@{}}Total\end{tabular}}} & \multicolumn{1}{l}{\textbf{Layer}} & \multicolumn{3}{c}{\textbf{Threshold}} \\ \hline
 & \multicolumn{1}{l}{} & \multicolumn{1}{l|}{0 samples} & \multicolumn{1}{l|}{100 samples} & \multicolumn{1}{l}{1000 samples} \\ \hline
Encodings & 1 & \multicolumn{1}{c|}{54,739} & \multicolumn{1}{c|}{0} & 0 \\
Split nodes & 1 & \multicolumn{1}{c|}{4,682} & \multicolumn{1}{c|}{0} & 0 \\ \hline
Encodings & 2 & \multicolumn{1}{c|}{52,265} & \multicolumn{1}{c|}{0} & 0 \\
Split nodes & 2 & \multicolumn{1}{c|}{1,341} & \multicolumn{1}{c|}{0} & 0 \\ \hline
Encodings & 3 & \multicolumn{1}{c|}{26,357} & \multicolumn{1}{c|}{27} & 0 \\
Split nodes & 3 & \multicolumn{1}{c|}{184} & \multicolumn{1}{c|}{6} & 0 \\ \hline
Encodings & 4 & \multicolumn{1}{c|}{5,236} & \multicolumn{1}{c|}{80} & 10 \\
Split nodes & 4 & \multicolumn{1}{c|}{33} & \multicolumn{1}{c|}{9} & 7 \\ \hline
Encodings & 5 & \multicolumn{1}{c|}{900} & \multicolumn{1}{c|}{62} & 14 \\
Split nodes & 5 & \multicolumn{1}{c|}{14} & \multicolumn{1}{c|}{9} & 9 \\ \hline
Encodings & 6 & \multicolumn{1}{c|}{342} & \multicolumn{1}{c|}{39} & 17 \\
Split nodes & 6 & \multicolumn{1}{c|}{10} & \multicolumn{1}{c|}{9} & 9 \\ \hline
Encodings & 7 & \multicolumn{1}{c|}{161} & \multicolumn{1}{c|}{32} & 13 \\
Split nodes & 7 & \multicolumn{1}{c|}{9} & \multicolumn{1}{c|}{9} & 9 \\ \hline
Encodings & 8 & \multicolumn{1}{c|}{138} & \multicolumn{1}{c|}{25} & 17 \\
Split nodes & 8 & \multicolumn{1}{c|}{9} & \multicolumn{1}{c|}{9} & 9 \\ \hline
Encodings & 9 & \multicolumn{1}{c|}{187} & \multicolumn{1}{c|}{34} & 13 \\
Split nodes & 9 & \multicolumn{1}{c|}{9} & \multicolumn{1}{c|}{9} & 9 \\ \hline
Encodings & 10 & \multicolumn{1}{c|}{196} & \multicolumn{1}{c|}{32} & 15 \\
Split nodes & 10 & \multicolumn{1}{c|}{9} & \multicolumn{1}{c|}{9} & 9
\end{tabular}
\end{table}

\begin{table}[H]
\caption{Average train and test accuracy of decision trees extracted at each layer of a MNIST-100 network with different thresholds applied.  Average is obtained over three different initialization seeds.}
\label{tab:accuracy_over_thresholds}
\centering
\begin{tabular}{llccc}
\hline
\textbf{Data Set} & \textbf{Layer} & \multicolumn{3}{c}{\textbf{Accuracy (\%)}} \\ \hline
 &  & \multicolumn{1}{l}{Threshold 0} & \multicolumn{1}{l}{Threshold 100} & \multicolumn{1}{l}{Threshold 1,000} \\ \hline
Train & 1 & 100.00 & 0.00 & 0.00 \\
Test & 1 & 89.60 & 0.00 & 0.00 \\ \hline
Train & 2 & 100.00 & 0.00 & 0.00 \\
Test & 2 & 95.67 & 0.00 & 0.00 \\ \hline
Train & 3 & 100.00 & 61.50 & 0.00 \\
Test & 3 & 97.47 & 61.05 & 0.00 \\ \hline
Train & 4 & 100.00 & 97.08 & 77.19 \\
Test & 4 & 97.77 & 94.79 & 75.58 \\ \hline
Train & 5 & 100.00 & 98.99 & 94.33 \\
Test & 5 & 97.85 & 96.41 & 92.08 \\ \hline
Train & 6 & 100.00 & 99.33 & 99.06 \\
Test & 6 & 97.84 & 96.88 & 96.65 \\ \hline
Train & 7 & 100.00 & 99.66 & 98.32 \\
Test & 7 & 97.85 & 97.29 & 95.67 \\ \hline
Train & 8 & 100.00 & 99.76 & 98.55 \\
Test & 8 & 97.85 & 97.41 & 96.03 \\ \hline
Train & 9 & 100.00 & 99.84 & 99.19 \\
Test & 9 & 97.78 & 97.48 & 96.87 \\ \hline
Train & 10 & 100.00 & 99.75 & 97.97 \\
Test & 10 & 97.81 & 97.28 & 95.75
\end{tabular}
\end{table}
\section{Decision Tree Visualization}
\label{sec:dt_viz}

We visualize the layerwise classification procedure for each class as a unique combination of node states, in the form of IF-THEN rules.  Green edges indicate that a node is required to be ON, while red indicates that a node must be OFF.  After each split node, we show two heatmaps, each consisting of the average of all samples for which the unique combination of node states are active: one indicating if the node is ON, and one indicating if the node is OFF.  Leaf nodes are shown as pink circles indicating the class.  

\begin{figure}[h]
        \centering
        \includegraphics[scale=0.10]{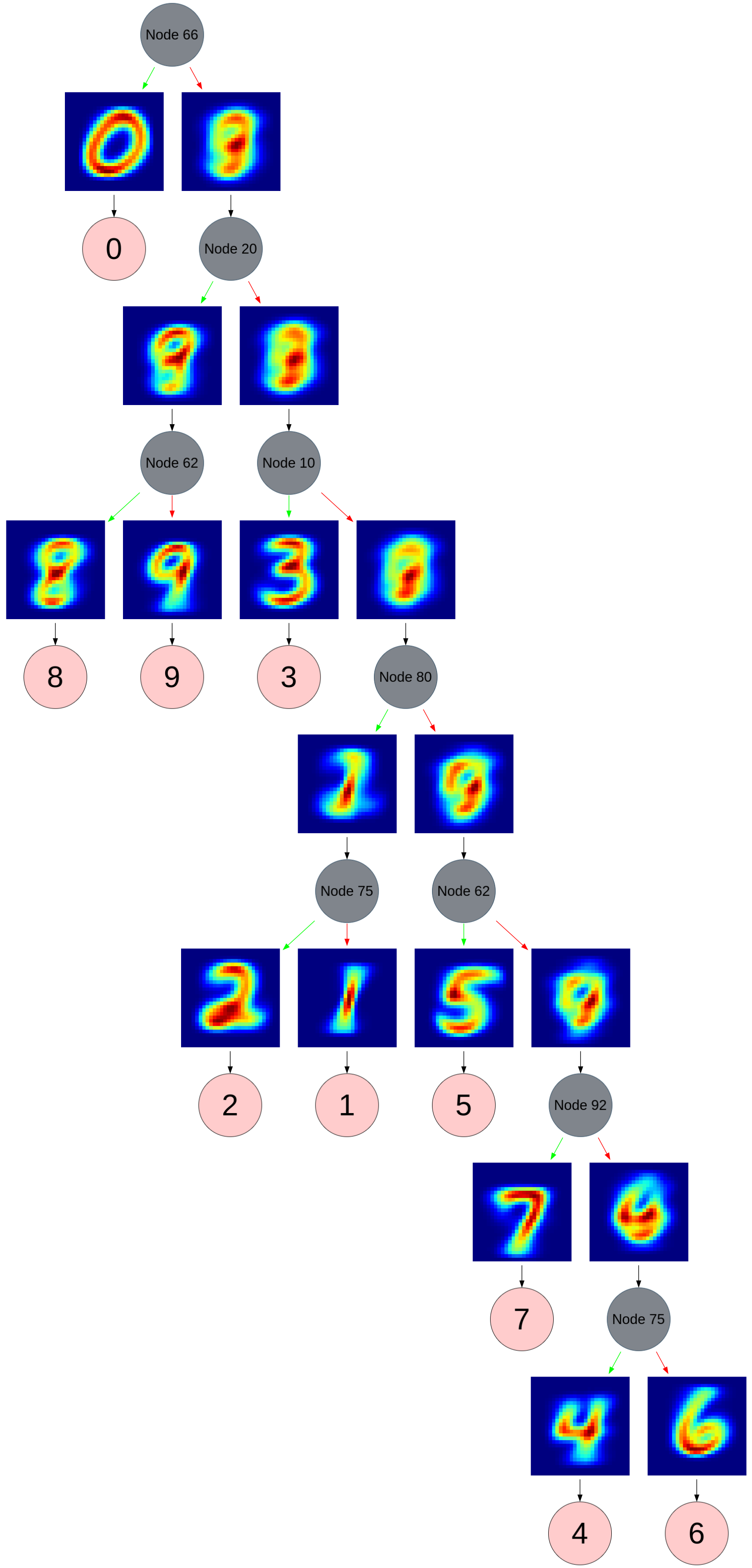}
        \caption{Visualization of a decision tree constructed from binary encodings found at the last hidden layer of a MNIST-100 network with 10 hidden layers.}
        \label{fig:dt_vizualization_layer_10_mnist}
\end{figure}

For example, consider Figure \ref{fig:dt_vizualization_layer_10_mnist}, where we show a visualization of a decision tree constructed from the binary encodings of the tenth (last) hidden layer of the MNIST-100 network, with no threshold applied.  The root node, node 66, shows a heatmap of all the samples for which the node is ON (green edge) and another for all the samples for the node is OFF (red edge).  The following split node, node 20, then shows heatmaps consisting of all the samples for which the node is ON or OFF AND for which the root node is OFF, etc.  Through showing the `cumulative' heatmaps at each decision point we are able to visualize how the samples of the training set are partitioned by each node's state.  

The visualization in Figure \ref{fig:dt_vizualization_layer_10_mnist} clearly shows how each node acts as a binary classifier, by partitioning the training set samples with its respective ON or OFF state. Certain nodes appear to only activate for samples of a single class, for example node 66 is clearly only active for samples from class 0.  However, several others appear to be a combination of different (but not all) classes, which allows for more effective partitioning of the training samples.

\begin{figure}[h]
        \centering
        \includegraphics[scale=0.08]{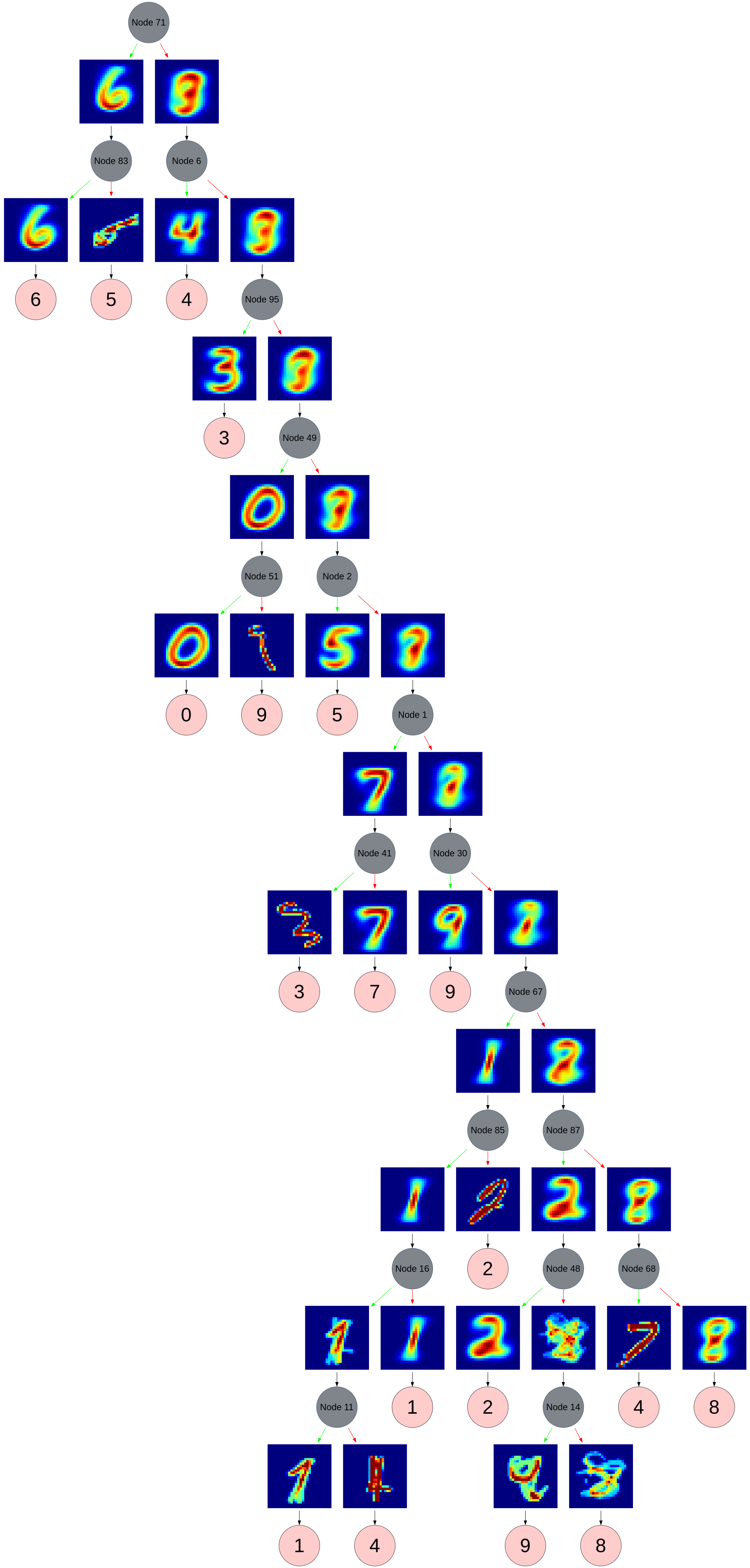}
        \caption{Visualization of a decision tree constructed from binary encodings found at the fifth hidden layer of a MNIST-100 network with 10 hidden layers.}
        \label{fig:dt_vizualization_layer_5_mnist}
\end{figure}

Figure \ref{fig:dt_vizualization_layer_5_mnist} shows the visualization of the decision tree derived from the fifth layer of the same network, also with no thresholding applied.  This visualization clearly shows the balance achievable through thresholding, as the classification process is much more complex due to encodings for idiosyncratic samples.  For example, the heatmaps shown for node 51 (OFF state) and 41 (ON state) are clearly for specific idiosyncratic samples that are not contained within their class majority encoding.  By applying a threshold of 100, the decision tree resembles that of Figure \ref{fig:dt_vizualization_layer_10_mnist}, and is much more concise (not shown due to space constraints). Depending on the complexity of the task and the goal of the analysis, a more concise or more detailed decision tree may be required.    

Finally, while these heatmaps are already informative, they can readily be swapped out with with any other saliency map visualization method such as SHAP~\cite{shap}, DeepLift~\cite{deep_lift}, or LRP~\cite{lrp}.  In this way, this method of decision tree induction is complementary to methods that provide interpretability for a single sample, and is able to provide a more global (layerwise) interpretation of a neural network.

It is also worth noting that these decision trees can only provide an approximation of the network's classification decisions, however, the similarity between the neural network's accuracy and that of the decision trees indicate that they closely mimic the neural network’s behaviour from an external perspective.

\section{Conclusion}

We have investigated the binary encodings formed by layers of a trained ReLU-activated MLP, and their subsequent behaviour on sample groupings.  Furthermore, we have demonstrated a straightforward approach to decision tree induction using these encodings, and how the resulting tree is able to illuminate the underlying layerwise decision making of a neural network.

Our main findings are summarised as follows:
\begin{itemize}
    \item Binary encodings can be used to derive decision trees, and these decision trees are able to provide excellent accuracy on both the train and test data sets.
    \item The decision trees can be used to visualize the decision-making process of a model. Demonstrated here with mean sample heatmaps, the heatmaps can be replaced with those produced by a variety of existing feature attribution methods.
    \item The threshold used to determine which encodings to remove controls the complexity of the tree (and visualization produced). Depending on the goal of the analysis, a more concise or more detailed tree can be produced. 
    \item Duplicate encodings occur when an encoding is shared by samples of more than a single class. Our analysis shows that these duplicates are caused, in part, by idiosyncratic samples which are visually very dissimilar to their class majority.
\end{itemize}
In terms of future work, we wish to first verify these results on more complex data sets before extending the process to additional attribution techniques and more complex architectures. We have already seen that activation functions that are not piecewise (such as sigmoid functions) can also be used to produce binary encodings, and that such encodings can be extracted from layers of more complex architectures. We aim to build on these observations to extend the current process to additional DNN architectures.  Finally, we would like to assess this method on a real-world application, in order to better understand its usefulness in practice, and the extent to which it can be used to supplement other interpretability methods.

\bibliographystyle{unsrt}  
\bibliography{references} 

\end{document}